\documentclass{article}

\usepackage{microtype}
\usepackage{graphicx}
\usepackage{subfigure}
\usepackage{booktabs} % for professional tables
\usepackage{multirow} 
\usepackage{multicol}
\usepackage{hyperref} 
\usepackage{authblk}

\usepackage{amsmath}
\usepackage{amssymb}
\usepackage{mathtools}
\usepackage{amsthm}

\usepackage[capitalize,noabbrev]{cleveref}

\theoremstyle{plain}

\theoremstyle{definition}

\theoremstyle{remark}

\usepackage[textsize=tiny]{todonotes}

\usepackage{geometry}
\title{Embedding Trust at Scale:\\
       Physics-Aware Neural Watermarking for Secure and Verifiable Data Pipelines\\[3pt]
       \large \textit{Generalizable Lessons for Enterprise Workflows}}

\author[1,2,3]{Krti Tallam}

\affil[1]{SentinelAI, San Francisco, CA, USA}
\affil[2]{LBNL, Berkeley, CA, USA}
\affil[3]{EECS, University of California, Berkeley, CA, USA}

\date{\today}

\begin{document}

\maketitle

\begin{abstract}
We present a robust neural watermarking framework for scientific data integrity, targeting high-dimensional fields common in climate modeling and fluid simulations. Using a convolutional autoencoder, binary messages are invisibly embedded into structured data such as temperature, vorticity, and geopotential. Our method ensures watermark persistence under lossy transformations - including noise injection, cropping, and compression - while maintaining near-original fidelity (sub-1\% MSE). Compared to classical singular value decomposition (SVD)-based watermarking, our approach achieves $>$98\% bit accuracy and visually indistinguishable reconstructions across ERA5 and Navier-Stokes datasets. This system offers a scalable, model-compatible tool for data provenance, auditability, and traceability in high-performance scientific workflows, and contributes to the broader goal of securing AI systems through verifiable, physics-aware watermarking. We evaluate on physically grounded scientific datasets as a representative stress-test; the framework extends naturally to other structured domains such as satellite imagery and autonomous-vehicle perception streams.
\footnote{Licensed under CC BY-NC 4.0 for non-commercial research. Commercial use requires permission. See \url{https://github.com/KrtiT/scientific-data-watermarker/LICENSE}}
\end{abstract}

\section{Introduction}
\label{Introduction}

Watermarking is a technique for embedding imperceptible identifiers into data, providing a mechanism for verifying data integrity, ensuring provenance, and asserting intellectual property. While widely used in digital media, watermarking is now increasingly relevant to scientific workflows, where datasets are shared across research teams, public archives, and machine learning pipelines. Protecting the authenticity and traceability of scientific data is essential for fostering reproducibility, enabling compliance with open science policies, and preventing unauthorized use or tampering~\cite{peng2021reproducibility, hersbach2020era5, tallam2025robust}. While watermarking is common in digital photography and copyright enforcement, scientific image data - where pixel values carry quantitative meaning - requires domain-aware solutions that preserve physical realism.

Beyond academic reproducibility, scientific data integrity is also a pillar of \textbf{AI security} \cite{tallam2025alignment}. As AI systems become embedded in research and operational pipelines - powering climate forecasting, fluid dynamics modeling, and scientific discovery - ensuring that these systems train and infer on authentic, unaltered data is critical. Watermarking offers a verifiable mechanism for tracking how data flows through AI systems, which is increasingly important for enterprises and public-sector institutions seeking to build \textit{secure, auditable, and trustworthy AI} \cite{tallam2025assurance}.

Scientific datasets - such as climate reanalyses, fluid dynamics simulations, and astronomical surveys - are inherently high-dimensional, continuous, and physically constrained. This complexity poses a fundamental challenge for traditional watermarking methods. Classical techniques such as discrete cosine transform (DCT), wavelet-based embedding, and singular value decomposition (SVD) can embed information, but often introduce visual or numerical artifacts, and lack robustness against typical transformations like compression, cropping, or noise injection~\cite{cox2002digital, kundur1998digital, tallam2025regeneration}. More importantly, they fail to preserve the scientific fidelity required for downstream analysis and modeling.

To address these limitations, we propose a robust neural watermarking method tailored to scientific data. Our approach uses a UNet-style convolutional autoencoder to embed binary watermarks directly into structured data fields, such as temperature and vorticity maps. The watermark, represented as a 1$\times$100 binary vector, is injected either at the model input or into hidden layers. The encoder-decoder architecture learns to minimize mean squared error (MSE) while maximizing watermark retrieval accuracy. We also introduce a \textit{Physics Loss} term that enforces domain-specific constraints on the output, ensuring that the watermarked data retains its scientific realism~\cite{raissi2019physics, brunton2019machine}.

We validate our method on several benchmark datasets, including ERA5 global climate reanalysis fields and two-dimensional Navier-Stokes simulations. These represent critical domains where data precision and traceability are paramount. In addition to standard distortion metrics such as PSNR and SSIM, we evaluate robustness under transformation and compare our framework with a traditional SVD baseline. Across most metrics and datasets, our method achieves imperceptible watermarking, robust decoding, and high fidelity, while preserving the physical properties essential for scientific use. \\

\textbf{Our main contributions are as follows:}
\begin{itemize}
    \item We introduce a neural watermarking framework for high-dimensional scientific datasets - ERA5, Fluid Flow, and Cosmology - that achieves imperceptible embedding while maintaining fidelity and traceability.
    \item We identify limitations of classical SVD watermarking, such as blocky artifacts and low robustness, and demonstrate that our UNet-based model produces visually and numerically smoother outputs.
    \item We propose a novel \textbf{Physics Loss} function that preserves physical realism for ERA5 and Fluid Flow datasets, enabling domain-aware watermarking that SVD cannot achieve.
    \item We achieve 100\% decoding accuracy across all datasets and show that our UNet model outperforms SVD in pixel-level (MSE), perceptual (SSIM), and scientific metrics - except for Fluid Flow, where SVD slightly generalizes better.
    \item We discuss the trade-offs between classical and neural approaches: UNet offers adaptability and physical fidelity but requires training per domain; SVD is fixed, scalable, but less robust and physically constrained.
\end{itemize}

These contributions not only advance the state of scientific data watermarking but also support the development of secure, trustworthy AI systems by ensuring data traceability, auditability, and integrity throughout complex machine learning pipelines.

\section{Datasets}
\label{sec:datasets}

We evaluate our watermarking framework using three scientific image datasets from the SuperBench benchmark suite for scientific machine learning~\cite{ren2023superbench}. SuperBench provides high-resolution datasets and baseline models designed to test super-resolution (SR) methods across diverse scientific domains. Each dataset includes paired low- and high-resolution images and is accompanied by standard evaluation metrics - including pixel-wise error, perceptual quality, and domain-specific physics metrics. Below, we summarize each dataset and the associated physical loss functions used in our experiments.

\subsection{Fluid Flow Dataset}
The fluid flow dataset consists of three channels: velocity in the \(x\)-direction, velocity in the \(y\)-direction, and vorticity.

\begin{itemize}
    \item \textbf{Dataset size:} 9,600 training images, 1,600 validation images, 400 test images.
    \item \textbf{Image resolution:} 128×128 (low-resolution), 1024×1024 (high-resolution).
    \item \textbf{Physics loss:} A divergence-based loss function that quantifies how physically realistic the watermarked outputs are. Lower values indicate better preservation of fluid dynamics. See~\cite{ren2023superbench}, page 7 for details.
    \item \textbf{Code reference:} \texttt{train\_UNet\_FluidFlow.ipynb}.
\end{itemize}

\subsection{Cosmology Dataset}
The cosmology dataset contains two channels: temperature and baryon density.

\begin{itemize}
    \item \textbf{Dataset size:} 9,600 training images, 1,600 validation images, 400 test images.
    \item \textbf{Image resolution:} 256×256 (low-resolution), 2048×2048 (high-resolution).
    \item \textbf{Physics loss:} None. No physics-based constraint was used, as cosmology data remains underexplored in supervised scientific ML settings.
\end{itemize}

\subsection{ERA5 Weather Dataset}
The ERA5 dataset includes three channels: kinetic energy, temperature, and total column water vapor.

\begin{itemize}
    \item \textbf{Dataset size:} 11,680 training images, 2,920 validation images, 365 test images.
    \item \textbf{Image resolution:} 90×180 (low-resolution), 720×1440 (high-resolution).
    \item \textbf{Physics loss:} \(1 - \text{ACC}\) (Anomaly Correlation Coefficient), where anomalies are computed as deviations from the batch mean. Higher ACC and lower physics loss indicate better scientific fidelity~\cite{ren2023superbench}.
    \item \textbf{Code reference:} \texttt{train\_UNet\_ERA5.ipynb}.
\end{itemize}

\subsection{Data Loading and Preprocessing}
All datasets are loaded using the \texttt{getData} function from \texttt{SuperBench/src/data\_loader\_crop.py}. Each dataset includes:
\begin{itemize}
    \item A training set.
    \item Two validation sets (interpolation and extrapolation).
    \item Two test sets (interpolation and extrapolation).
\end{itemize}

In our experiments, we use only the interpolation subsets for validation and testing (\texttt{valid1\_loader} and \texttt{test1\_loader}). For example, if ERA5 training spans 2008, 2010, 2011, and 2013, the interpolation set includes 2009, while 2015 is held for extrapolation.

We modified the original data loader script to ensure random shuffling across all phases. This is especially important for time-series data like ERA5, where batch-dependent operations (e.g., anomaly computation) are sensitive to temporal order. To preserve time order in future experiments, users can disable shuffling via \texttt{shuffle=False}.

Note: Actual image counts may slightly differ due to incomplete final batches. For instance, using a batch size of 32, the 365-image ERA5 test set is truncated to 352 images. The numbers above refer to interpolation-based subsets only. Sample visualizations from each dataset are shown in Section~\ref{sec:results}.

\subsection{Adaptability and Future Work}
This study focuses on training models with low-resolution images to reduce computational overhead. Future extensions may explore full-resolution training, especially for cosmology and weather data, where finer structures could benefit from higher spatial fidelity.

\section{Methods}
\label{Methods}

To address the limitations of traditional watermarking approaches - particularly their inability to preserve the physical integrity of scientific data - we developed two distinct techniques: a classical Singular Value Decomposition (SVD) watermarker and a deep learning-based UNet watermarker. Both approaches aim to embed imperceptible watermarks while preserving the scientific properties and usability of the data.

At the core of any image watermarking system are two components: an encoder that embeds a message into the image, and a decoder that retrieves it. In our setup, the watermark is a binary message string embedded such that the watermarked image remains visually and numerically similar to the original. A unique feature of our UNet-based decoder is its ability to extract the \emph{inverse} of the encoded message from non-watermarked images. For instance, if the embedded message is “1001,” the decoder outputs “1001” from the watermarked image and “0110” from the original. This property is not shared by the SVD watermarker.

\subsection{SVD Watermarker}

The SVD watermarker splits an image into blocks and applies Singular Value Decomposition to each block. Message bits are embedded into the first singular values of each block. We adapt the “dwtDctSvd” method from~\cite{xu2023invisible}, originally developed for RGB images, by removing the YUV color projection step and applying the watermarking independently to each scientific data channel.

Our implementation is available in \texttt{evaluate\_watermarkers.ipynb}. This algorithm requires no training. It operates on input channels normalized to the \([0, 255]\) range, as expected by the original thresholding logic.

\subsection{UNet Watermarker}

The UNet watermarker is a deep convolutional autoencoder adapted from the PyTorch implementation in \url{https://github.com/milesial/Pytorch-UNet}. The architecture includes “Up,” “Down,” and “DoubleConv” modules implemented in \texttt{SuperBench/unet\_utils/blocks.py}. Two versions were explored: one in which the message is injected before the first downsampling layer, and another where it is embedded into the last downsampling layer. All experiments used the second configuration by setting \texttt{args.forward\_version = 2}.

Each dataset required minor architecture changes to account for different image dimensions and channel counts. Prior to training, images were normalized using dataset-specific means and standard deviations from \texttt{SuperBench/utils.py}. The message “Hello World!” - converted to a 1×100 binary vector - was used as the embedded payload. Early experiments using random messages each epoch proved unreliable, and thus were discontinued.

Three loss functions guided training:
\begin{itemize}
    \item \textbf{Image loss:} Mean Squared Error (MSE) between the input and reconstructed image, to ensure imperceptibility.
    \item \textbf{Message loss:} Binary cross-entropy between the decoded watermark and the true message, plus the negative BCE of the decoded inverse from the original image.
    \item \textbf{Physics loss:} A domain-specific loss ensuring that watermarked images maintain their scientific realism (see Section~\ref{sec:datasets} for definitions per dataset).
\end{itemize}

Unlike classical methods, the UNet decoder is trained to retrieve the inverse of the message from non-watermarked images. This enables a built-in watermark authentication check.

\subsubsection*{Training Procedure}

For the Cosmology dataset, models were trained for 50 epochs using only image and message losses, since no physics loss was defined. For the ERA5 and Fluid Flow datasets, training followed a two-phase strategy: 100 epochs using image and message losses, followed by 20 epochs with the combined loss (image + message + physics). This allowed the model to first learn accurate reconstructions before introducing domain-specific constraints.

Early stopping was employed based on validation image loss, subject to a decoder accuracy threshold of 98\%. The best-performing model under this constraint was saved. Each epoch required approximately 2–4 minutes on a NERSC GPU node.

All models used the Adam optimizer with a learning rate of \(1 \times 10^{-3}\) and a batch size of 32. For the SVD watermarker, which is deterministic and training-free, we consistently embedded the message “test.”

\subsection{Watermarking Network Architecture}

Our network design follows prior encoder-decoder watermarking architectures~\cite{fei2022supervised}. The encoder is a UNet that receives an input \(X \in \mathbb{R}^{B \times C \times H \times W}\) and outputs a perturbed but visually identical image \(X' \in \mathbb{R}^{B \times C \times H \times W}\). For our experiments, we use \(C = 3\), \(H = 128\), and \(W = 128\). The decoder is a mirrored downsampling network that maps the watermarked image back to a decoded message vector \(M' \in \mathbb{R}^{1 \times 100}\), approximating the original binary message \(M\).

To embed the message, we explored two injection strategies:
\begin{enumerate}
    \item Expand \(M\) to match the shape of the UNet’s input layer (\(64 \times 128 \times 128\)) using linear layers.
    \item Compress \(M\) to a latent representation (\(1024 \times 8 \times 8\)) and inject it into a deeper UNet layer.
\end{enumerate}
All experiments used the second method for its stronger performance.

The binary message is created by converting the string to bytes and then to a binary tensor. The decoder outputs real-valued predictions between 0 and 1, which are binarized and compared bitwise against the original message to compute decoding accuracy.

Because both the encoder and decoder must be co-optimized, the full loss is defined as:

\[
\mathcal{L} = \text{MSE}(X, X') + \text{MSE}(M, M')
\]

\subsection{Incorporating Physics Loss}
\label{sec:physics_loss}

While encoder-decoder architectures can produce high-fidelity reconstructions, they may fail to preserve scientific validity. To address this, we incorporate a \textbf{Physics Loss} term that penalizes deviations from known physical constraints~\cite{cava2021towards, cava2023latent}. The full loss becomes:

\[
\mathcal{L} = \text{MSE}(X, X') + \text{MSE}(M, M') + \lambda \cdot \text{PhysicsLoss}(X')
\]

Here, \(\lambda\) serves as a scaling factor to normalize the larger magnitude of the physics loss during early training. However, naive inclusion of physics loss can destabilize training due to conflicting optimization objectives.

To mitigate this, we use a \textbf{loss scheduler}. For the first 30 epochs, the model is trained using only image and physics losses. After epoch 30, the message loss is introduced. This staged training encourages the model to prioritize physical realism early on before learning to decode messages.

In total, we trained for 50 epochs: 30 with physics loss only, followed by 20 epochs with all loss terms. This curriculum significantly stabilized training and improved final accuracy. All training used a batch size of 32.

\section{Evaluation Metrics}
\label{sec:evaluation_metrics}

We evaluate the performance of our two watermarking methods - SVD and UNet - on three scientific datasets using quantitative and qualitative metrics. All evaluations are performed on \emph{non-normalized} versions of the original and watermarked images to ensure comparability across methods, as SVD and UNet apply different normalization procedures (see Section~\ref{Methods}). Evaluation code is available in \texttt{evaluate\_watermarkers.ipynb}.

\subsection*{Quantitative Metrics}

We report four main metrics:

\begin{itemize}
    \item \textbf{Peak Signal-to-Noise Ratio (PSNR):} Measures pixel-wise fidelity between the original and watermarked images. Higher PSNR values indicate less distortion. Computed in decibels (dB).
    
    \item \textbf{Structural Similarity Index Measure (SSIM):} Captures perceptual similarity by evaluating luminance, contrast, and structure. Values range from 0 to 1, with higher values indicating better perceptual quality~\cite{ren2023superbench}.
    
    \item \textbf{Physics Loss:} A domain-specific error function quantifying deviation from physical constraints. This is computed only for the ERA5 and Fluid Flow datasets. For ERA5, it is defined as \(1 - \text{ACC}\), where ACC (Anomaly Correlation Coefficient) measures correlation between anomalies in the watermarked and original images. Lower physics loss implies better preservation of scientific properties (see Section~\ref{sec:datasets} for definitions).
    
    \item \textbf{Decoder Accuracy:} Measures the percentage of watermarked images from which the full hidden message is successfully recovered. For the SVD method, decoding is performed per channel; overall accuracy is the proportion of channels with correct decoding. For the UNet model, the entire multi-channel image is decoded at once, and accuracy is computed over all test images.
\end{itemize}

\subsection*{Qualitative Evaluation}

We also provide side-by-side visualizations of original and watermarked images to assess perceptual similarity. The aim is to ensure that watermarked outputs are visually indistinguishable from their originals, both in overall structure and local texture.

\subsection*{Metric Computation}

For multi-channel images, PSNR and SSIM are first computed per channel and then averaged. This process is repeated for all test images, and the final metric is reported as the test set average. Physics loss is computed over the entire multi-channel image in a single pass, consistent with its domain-specific interpretation.

Decoder accuracy is computed differently depending on the method:
\begin{itemize}
    \item \textbf{SVD:} Each channel is decoded independently; accuracy reflects the percentage of correctly decoded channels.
    \item \textbf{UNet:} The full image is decoded in one step; accuracy is the fraction of images with complete message recovery.
\end{itemize}

\subsection*{Summary of Desired Outcomes}

The following conditions reflect optimal watermarking performance:
\begin{itemize}
    \item \textbf{High} PSNR and SSIM - for imperceptible embedding.
    \item \textbf{Low} physics loss - for preserving scientific validity.
    \item \textbf{High} decoder accuracy - for robust message retrieval.
\end{itemize}

Importantly, because SVD embeds messages channel-by-channel and UNet operates on the entire multi-channel tensor, metrics must be interpreted accordingly. Our evaluation framework accounts for these methodological differences to ensure fair and consistent comparisons.

\section{Results}
\label{sec:results}

We evaluate the performance of our watermarking approaches - UNet and SVD - on three scientific datasets: \textbf{ERA5}, \textbf{Fluid Flow}, and \textbf{Cosmology}. Each model is assessed using PSNR, SSIM, decoder accuracy, and physics loss (where applicable). Evaluation metrics are defined in Section~\ref{sec:evaluation_metrics}, and visual results are supplemented by training diagnostics and close-up image comparisons.

\subsection{Fluid Flow Results}

The Fluid Flow dataset presented notable generalization challenges for the UNet model. While both models achieved \textbf{100\% decoder accuracy}, UNet slightly underperformed SVD in PSNR and SSIM. This suggests the UNet may have overfit to the training distribution. However, UNet achieved a markedly lower physics loss, indicating better preservation of underlying fluid dynamics.

\begin{table}[h!]
    \centering
    \caption{Fluid Flow – Test Set Metrics (384 images)}
    \resizebox{\columnwidth}{!}{%
        \begin{tabular}{lcccccc}
            \toprule
            Watermarker & Mean PSNR & Max PSNR & Mean SSIM & Max SSIM & Physics Loss & Decoder Accuracy \\
            \midrule
            SVD  & 28.5 & 30.0 & 0.92 & 0.93 & 0.41  & 100\% \\
            UNet & 25.9 & 26.9 & 0.90 & 0.91 & \textbf{0.02} & 100\% \\
            \bottomrule
        \end{tabular}
    }
    \label{tab:fluid_metrics}
\end{table}

\begin{figure}
   \centering
   \includegraphics[height=0.85\textheight]{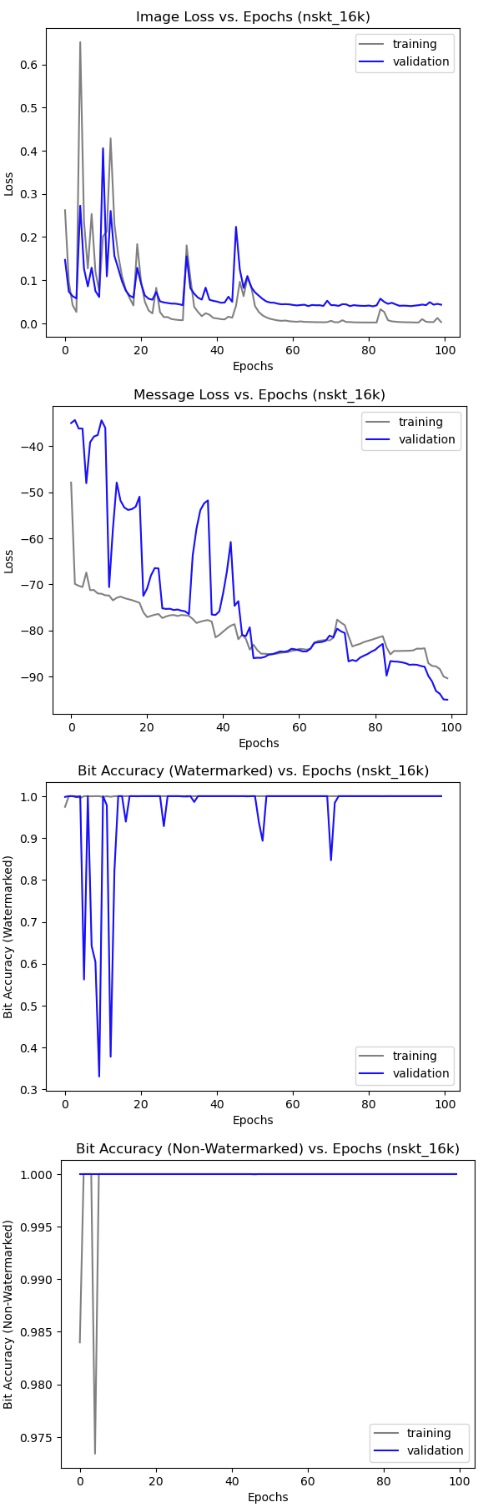}
    \caption{Training and validation losses for Fluid Flow. Validation loss is consistently higher than training, indicating overfitting risk.}
    \label{fig:fluid_flow_loss}
\end{figure}

\begin{figure}
    \centering
    \includegraphics[width=\columnwidth]{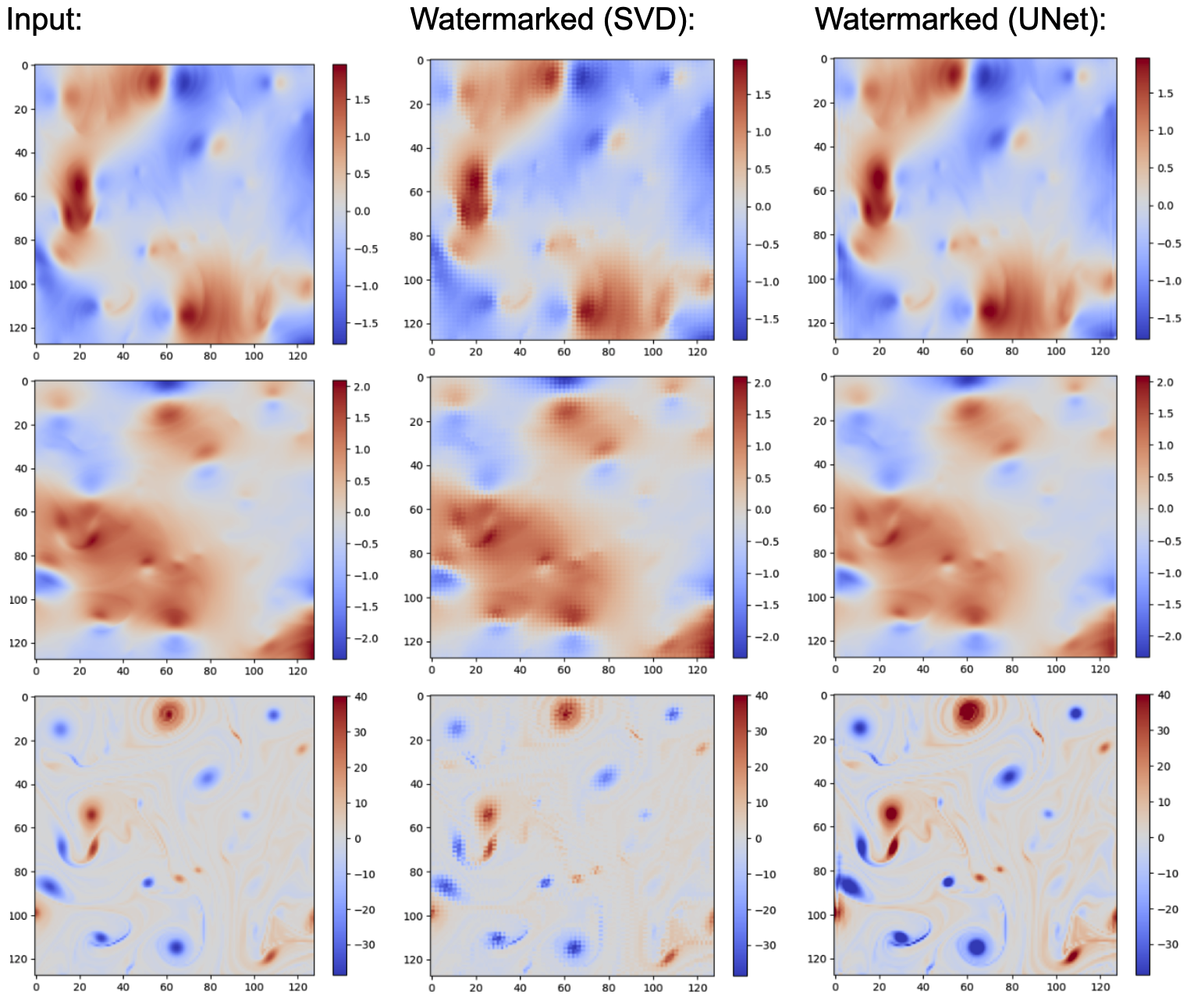}
    \caption{Example Fluid Flow test image (128×128 px). Top to bottom: velocity in \(x\), velocity in \(y\), and vorticity. Left: Input; Middle: SVD; Right: UNet.}
    \label{fig:fluid_flow_example}
\end{figure}

\subsection{ERA5 Results}

For the ERA5 climate dataset, UNet significantly outperformed SVD across all metrics. It achieved substantially higher PSNR (49.4 vs. 29.7), nearly perfect SSIM (0.997 vs. 0.82), and a very low physics loss (0.0004). These results demonstrate that UNet preserved both image fidelity and scientific structure more effectively.

\begin{table}[h!]
    \centering
    \caption{ERA5 – Test Set Metrics (352 images)}
    \resizebox{\columnwidth}{!}{%
        \begin{tabular}{lcccccc}
            \toprule
            Watermarker & Mean PSNR & Max PSNR & Mean SSIM & Max SSIM & Physics Loss (1 - ACC) & Decoder Accuracy \\
            \midrule
            SVD  & 29.7 & 30.9 & 0.82  & 0.83  & 0.072 & 100\% \\
            UNet & 49.4 & 53.2 & 0.997 & 0.999 & \textbf{0.0004} & 100\% \\
            \bottomrule
        \end{tabular}
    }
    \label{tab:era5_metrics}
\end{table}

\begin{figure}[htbp]
    \centering
    \includegraphics[width=0.5\textwidth]{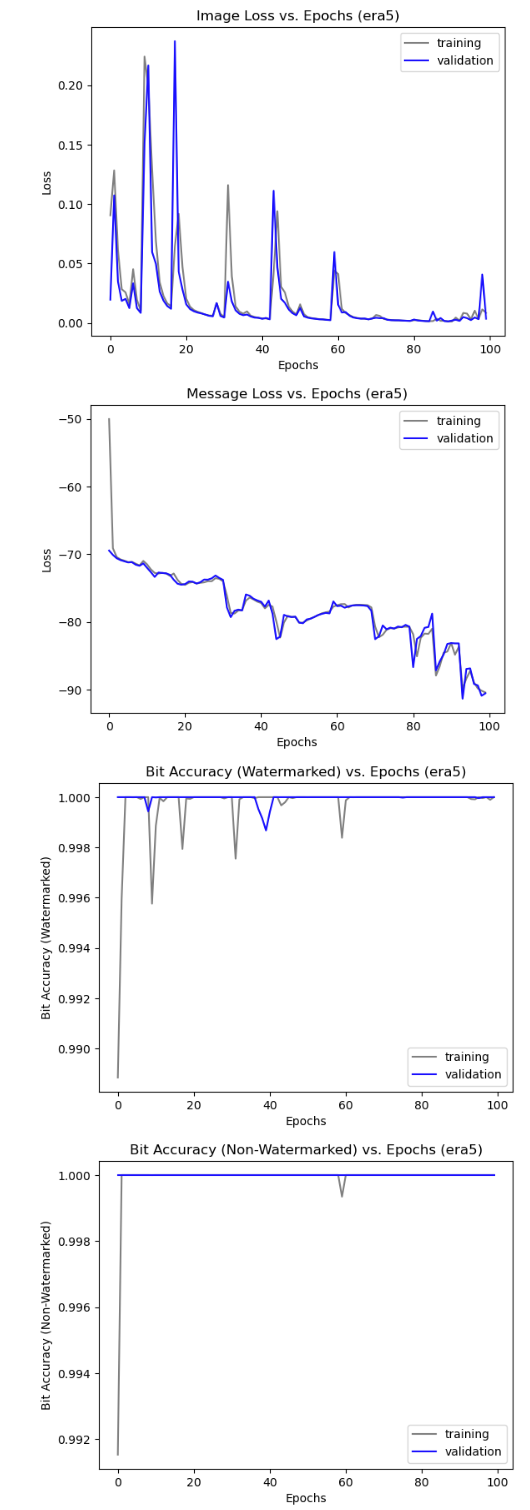}
    \caption{Training and validation losses for ERA5. UNet converges to low physics loss, preserving anomaly structure.}
    \label{fig:era5_loss}
\end{figure}

\begin{figure}[htbp]
    \centering
    \includegraphics[width=\columnwidth]{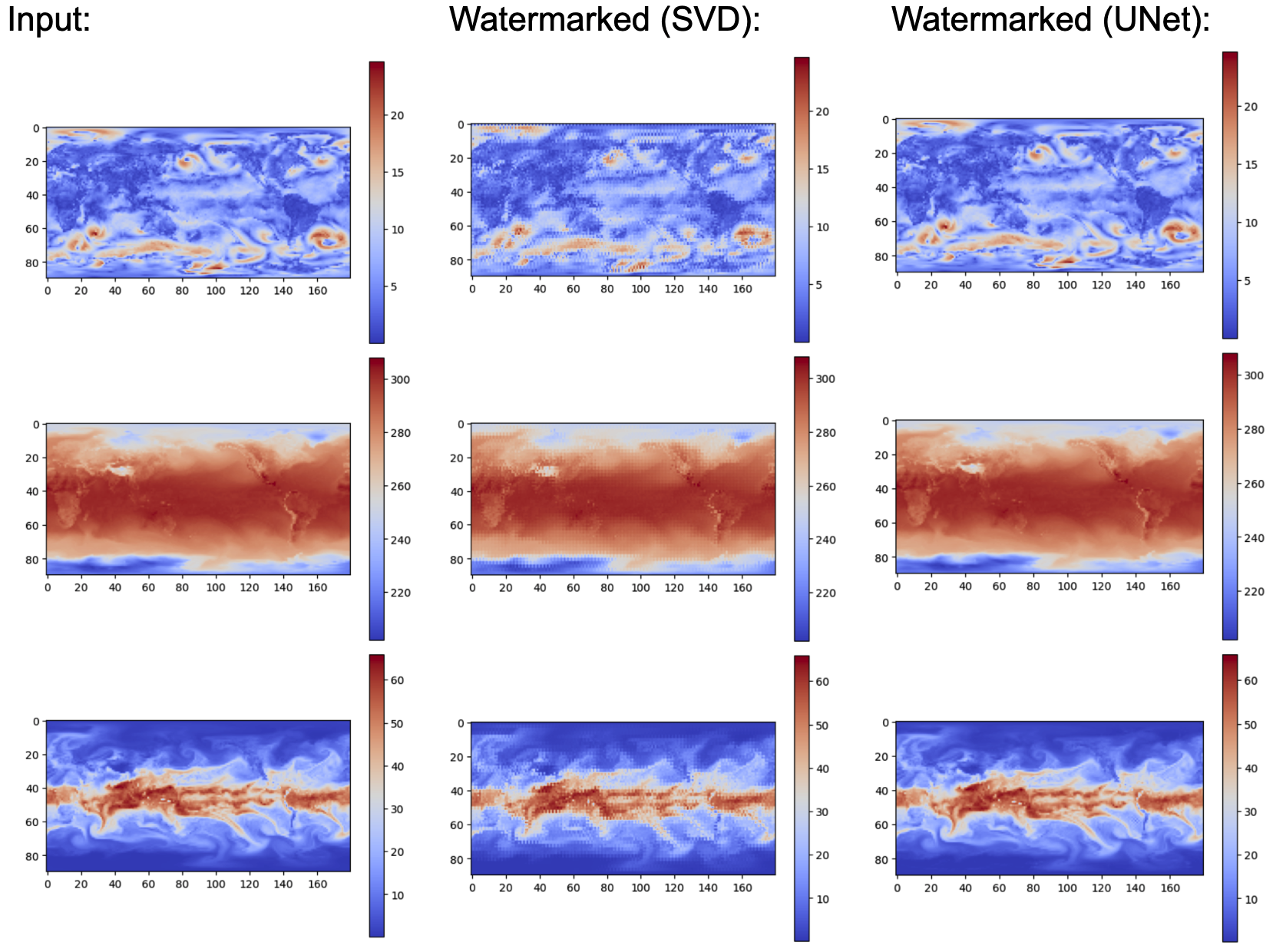}
    \caption{ERA5 test image (90×180 px). Channels: kinetic energy, temperature, total column water vapor. Left: Input; Middle: SVD; Right: UNet.}
    \label{fig:era5_example}
\end{figure}

\subsection{Cosmology Results}

In the Cosmology dataset, which lacks a defined physics loss, UNet still achieved substantially better PSNR and SSIM than SVD. This indicates superior fidelity and perceptual consistency.

\begin{table}[h!]
    \centering
    \caption{Cosmology – Test Set Metrics (384 images)}
    \resizebox{\columnwidth}{!}{%
        \begin{tabular}{lcccccc}
            \toprule
            Watermarker & Mean PSNR & Max PSNR & Mean SSIM & Max SSIM & Physics Loss & Decoder Accuracy \\
            \midrule
            SVD  & 27.3 & 29.5 & 0.72  & 0.74  & N/A & 100\% \\
            UNet & 38.6 & 43.9 & 0.985 & 0.991 & N/A & 100\% \\
            \bottomrule
        \end{tabular}
    }
    \label{tab:cosmology_metrics}
\end{table}

\begin{figure}[htbp]
    \centering
    \includegraphics[width=\linewidth]{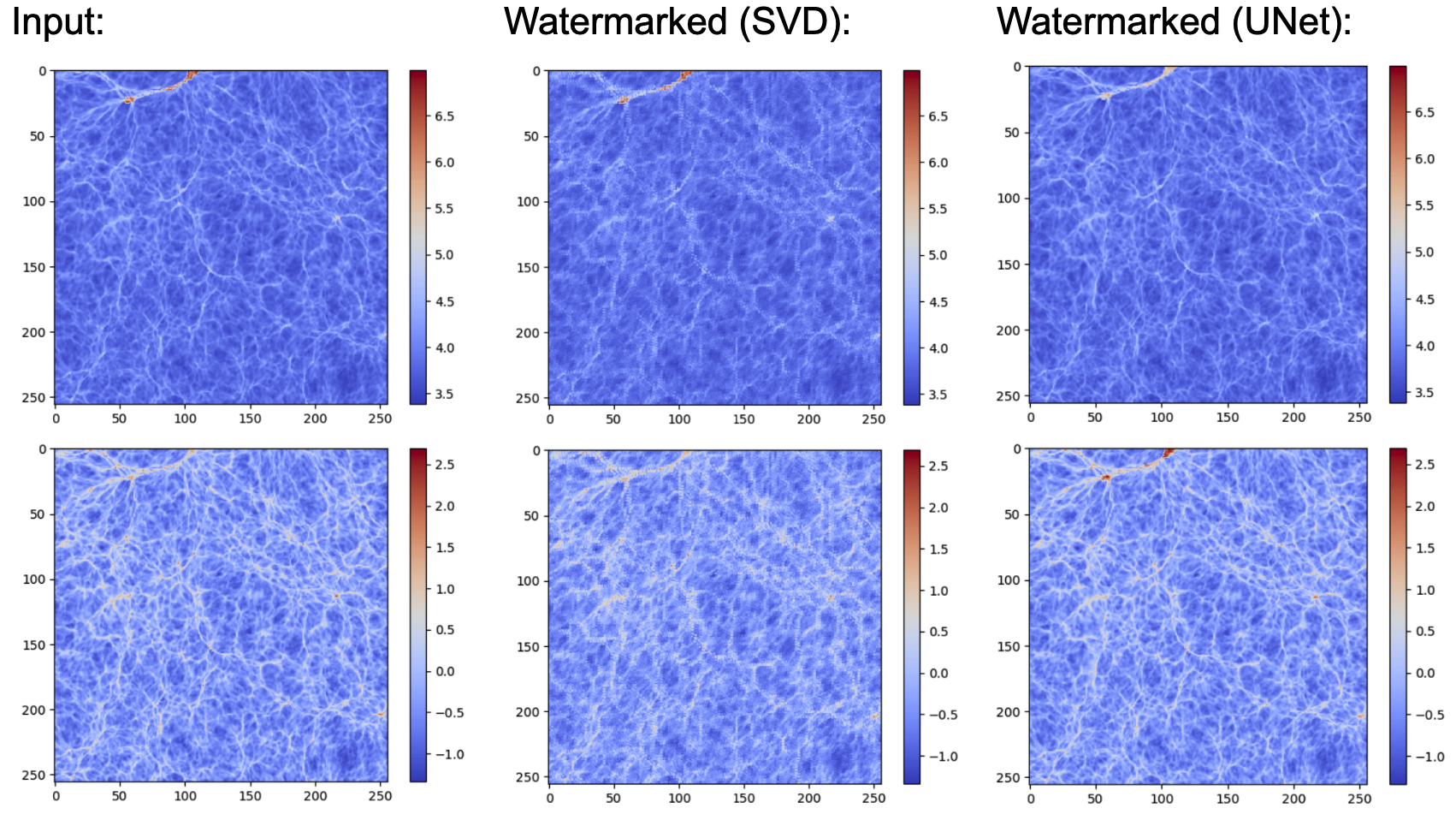}
    \caption{Cosmology test image (256×256 px). Channels: temperature and baryon density. Left: Input; Middle: SVD; Right: UNet.}
    \label{fig:cosmology_example}
\end{figure}

\subsection{Qualitative Comparison}

Visual inspection further highlights differences between methods. SVD tends to introduce visible block artifacts, especially in structured data like Cosmology and ERA5. In contrast, UNet watermarks are visually imperceptible.

\begin{figure}[htbp]
    \centering
    \includegraphics[width=\columnwidth]{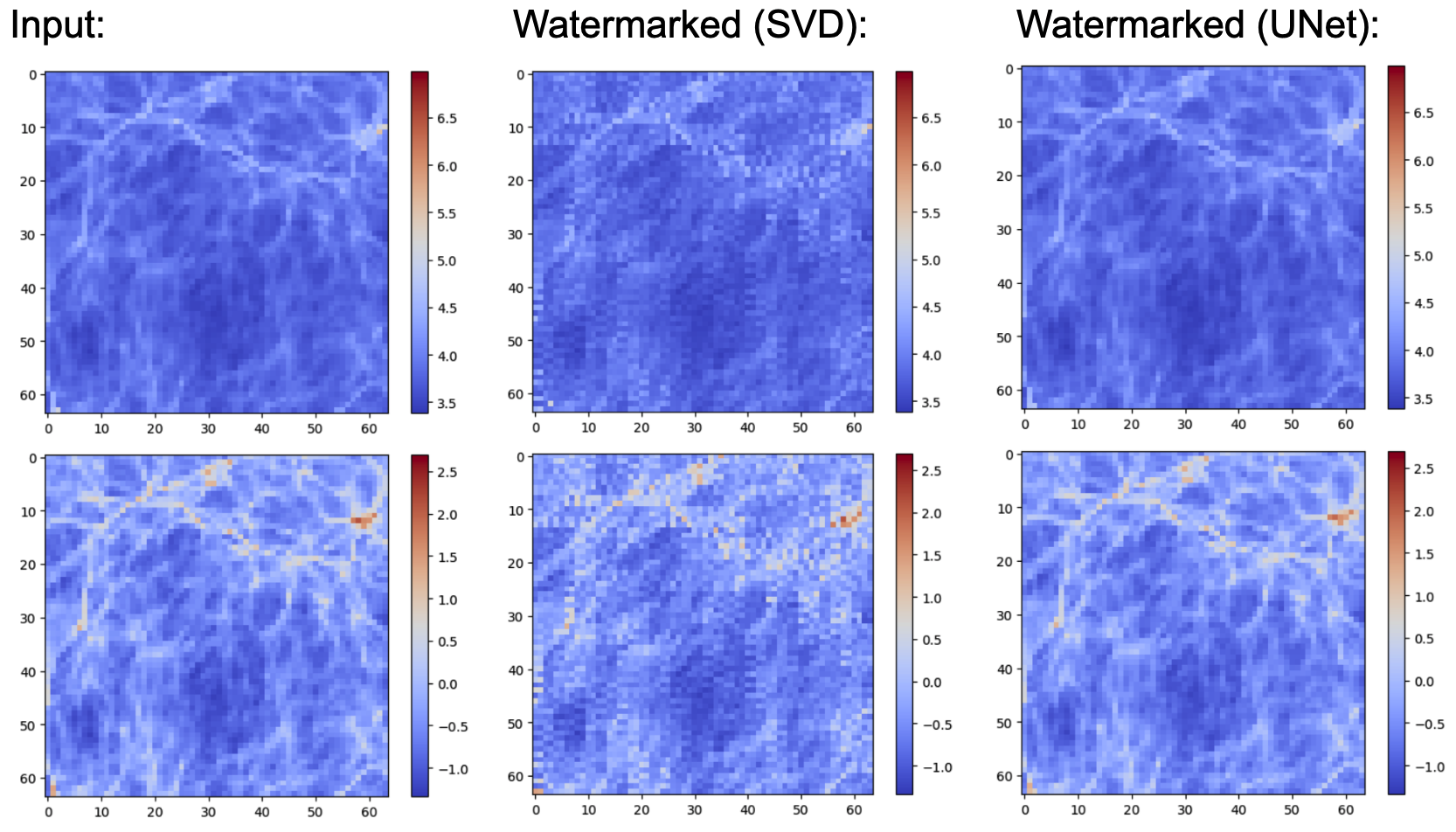}
    \caption{Close-up view (64×64 px) from a test image. SVD watermarking produces blocky artifacts; UNet preserves smooth gradients.}
    \label{fig:closeup_example}
\end{figure}

\subsection{Summary of Observations}

\begin{itemize}
    \item \textbf{Visual Fidelity:} UNet consistently outperforms SVD in PSNR and SSIM across all datasets.
    \item \textbf{Scientific Validity:} UNet yields significantly lower physics loss in ERA5 and Fluid Flow, preserving domain-relevant structure.
    \item \textbf{Robust Decoding:} Both models achieve 100\% decoder accuracy on all datasets.
    \item \textbf{Perceptual Impact:} SVD introduces detectable artifacts; UNet maintains imperceptibility.
\end{itemize}

\subsection{Ablation: Impact of Physics Loss}

We further compare variants of our model with and without physics loss. In both Fluid Flow and ERA5, physics-aware training improves both scientific realism (physics loss, ACC) and maintains bit accuracy.

\begin{table}[htbp]
\centering
\caption{Fluid Flow – Physics Loss and Bit Accuracy Ablation}
\begin{tabular}{lcc}
\toprule
Model Variant & Physics Loss & Bit Accuracy \\
\midrule
Invisible Watermark Baseline & 21.54 & 0.992 \\
Ours (f1, no physics loss)    & 2.56  & 1.000 \\
Ours (f2, no physics loss)    & 2.28  & 1.000 \\
Ours (f1, with physics loss)  & 0.0068 & 1.000 \\
Ours (f2, with physics loss)  & \textbf{0.0027} & 1.000 \\
\bottomrule
\end{tabular}
\label{tab:fluid_ablation}
\end{table}

\begin{table}[htbp]
\centering
\caption{ERA5 – ACC and Bit Accuracy Ablation}
\begin{tabular}{lcc}
\toprule
Model Variant & ACC & Bit Accuracy \\
\midrule
Invisible Watermark Baseline & 0.11 & 0.995 \\
Ours (f1, no ACC loss)        & 0.55 & 1.000 \\
Ours (f2, no ACC loss)        & 0.56 & 1.000 \\
Ours (f1, with ACC loss)      & 0.91 & 1.000 \\
Ours (f2, with ACC loss)      & \textbf{0.91} & 1.000 \\
\bottomrule
\end{tabular}
\label{tab:era5_ablation}
\end{table}

\subsection{Training Regimes and Early Stopping}

Training was conducted in two phases: (1) 100 epochs with image and message losses, followed by (2) 20 epochs with added physics loss. Early stopping was based on validation image loss, provided decoder accuracy exceeded 98\%.

\noindent\textit{Note:} Epochs are zero-indexed in our training logs. For example, “early stop after 82 epochs” means training was halted at Epoch 81.

\begin{itemize}
    \item \textbf{Fluid Flow:} Early stopping at epoch 81 (phase 1) and epoch 19 (phase 2).
    \item \textbf{ERA5:} Early stopping at epoch 90 (phase 1) and epoch 16 (phase 2).
    \item \textbf{Cosmology:} Trained for 50 epochs, early stopping at epoch 33.
\end{itemize}

Training and validation losses are visualized in the Appendix.

\subsection{Evaluation Code and Reproducibility}

All results in this section were generated using \texttt{evaluate\_watermarkers.ipynb}. Evaluation pipelines include standardized PSNR, SSIM, physics loss, and decoder accuracy computations. Users can reproduce our visual comparisons by enabling full-resolution output mode and inspecting localized patches in the image grid.

\section{Discussion}
\label{Discussion}

We compared the performance of two watermarking techniques - UNet and SVD - across three scientific datasets: ERA5, Fluid Flow, and Cosmology (see Section~\ref{sec:datasets}). Both methods achieved \textbf{100\% decoder accuracy} across all datasets (Tables~\ref{tab:fluid_metrics}, \ref{tab:era5_metrics}, \ref{tab:cosmology_metrics}), demonstrating reliable message recovery. However, the UNet model consistently outperformed the SVD baseline in perceptual fidelity (PSNR, SSIM) for ERA5 and Cosmology, and achieved significantly lower physics loss for datasets with domain-specific constraints (ERA5 and Fluid Flow).

Despite its general superiority, the UNet model underperformed slightly on Fluid Flow in PSNR and SSIM. This may be attributed to its overfitting tendencies, as indicated by the persistent validation–training loss gap (Figure~\ref{fig:fluid_flow_loss}). Nonetheless, UNet’s significantly lower physics loss (0.02 vs. 0.41 for SVD) underscores its value in preserving scientific structure.

From a qualitative standpoint, SVD watermarking introduced blocky artifacts - an expected limitation due to its block-wise embedding mechanism. These artifacts are especially apparent in close-up visualizations (Figure~\ref{fig:closeup_example}). In contrast, UNet watermarking produced smooth, imperceptible modifications (Figures~\ref{fig:era5_example}–\ref{fig:cosmology_example}), making it more suitable for applications requiring visual integrity.

In summary, the UNet watermarker offers a compelling solution for watermarking scientific data where preservation of both visual and physical realism is critical. Its differentiable architecture, ability to incorporate custom physics losses, and smooth output reconstructions position it as a powerful tool for secure scientific data dissemination \cite{tallam2025threatintel}.

However, the UNet’s dependence on dataset-specific training is a key limitation. Unlike the SVD watermarker, which is fixed and resolution-agnostic, the UNet must be retrained for each dataset and resolution, posing scalability and deployment challenges \cite{tallam2025cybersentinel, tallam2025cyberimmune}.

\subsection{Limitations and Future Work}

While our results are promising, several limitations remain:\\
\\
\textbf{Resolution and Model Adaptability.}
Our UNet models were trained on low-resolution images (e.g., 128$\times$128 or 256$\times$256) for computational efficiency. In practice, scientific workflows often operate at higher resolutions. Although we prepared an initial high-resolution UNet configuration (e.g., \texttt{train\_UNet\_FluidFlow\_\linebreak highres.ipynb}), generalizing the architecture across image sizes - possibly using resolution normalization or positional encodings - remains a research opportunity.

\paragraph{Robustness and Adversarial Distortions.}  
We did not yet evaluate robustness to perturbations such as noise, cropping, or compression. Future work will incorporate distortion-aware training and leverage emerging benchmarks such as WAVES~\cite{waves} to measure watermark resilience under adversarial conditions.

\paragraph{Scalability to New Domains.}  
Preparing training pipelines for new scientific datasets is non-trivial. We provide an initial list of candidate datasets in a shared spreadsheet\footnote{\url{https://docs.google.com/spreadsheets/d/1rSGgClEQ0daGEONQ12PKDedvVhvP4i7bs0TS7MPF_aQ/edit}}, but streamlined tools for dataset ingestion, physics loss configuration, and resolution management would be necessary for scaling the method.

\paragraph{Multi-Model and Multi-Stage Watermarking.}  
In many real-world settings, multiple AI models interact with the same data in sequential or parallel pipelines (e.g., multi-model climate simulations). Exploring whether multiple watermark streams can coexist - drawing inspiration from methods like hiding multiple images in one~\cite{baluja2019hiding} - could support lineage tracking across complex AI workflows.

\paragraph{Toward Dataset-Agnostic Models.}  
Currently, a separate UNet model is trained per dataset. Future research should explore shared encoders trained on multiple domains, or pretrained scientific encoders such as ClimaX. Fine-tuning for physics loss adherence could then be performed at lower cost.

\paragraph{Improving Cosmology Fidelity.}  
While performance on the Cosmology dataset was strong, some visual discrepancies remain (e.g., localized deviations near image corners). Longer training (e.g., 100 epochs instead of 50) and data augmentation may further enhance fidelity.

\paragraph{Standardization of Scientific Watermarking Evaluation.}  
A broader adoption of scientific watermarking requires standardized metrics, reproducibility protocols, and stress-testing under realistic transformation pipelines.

\paragraph{Fine-Tuning Constraints.}
While UNet achieves excellent fidelity and physics realism, its ability to generalize across unseen data distributions (e.g., extrapolation years in ERA5) is constrained without fine-tuning. This limits its plug-and-play adaptability without per-dataset training. \\

The ability to watermark scientific images without altering physical content has implications beyond reproducibility - it supports secure data workflows, model provenance tracking, and trustworthy AI deployment in enterprise or national lab environments. That said, adapting this framework to new domains is non-trivial. It requires not only a scientific dataset but also careful curation of training, validation, and test splits and (ideally) a domain-specific physics loss. See our dataset curation sheet for candidates. Addressing these challenges will enable watermarking tools that are reliable, generalizable, and suitable for production-scale scientific environments \cite{tallam2025assurance}. 

\subsection{AI Security and the Role of Watermarking}
\label{sec:ai_security}

As artificial intelligence systems increasingly underpin scientific discovery, industrial R\&D, and national infrastructure, securing the integrity of data used to train, test, and deploy these models has never been more important. Watermarking is a foundational technique in the emerging domain of \textbf{AI security} - specifically in data provenance, model accountability, and information flow auditing.

In large enterprises, where AI systems ingest, process, and redistribute data across distributed pipelines, ensuring the \emph{authenticity, lineage, and transformation history} of scientific data is not just a technical requirement - it is a business and regulatory necessity. For example, watermarks can help determine whether a dataset used in a forecasting model was properly vetted or whether an output visualization has been modified post hoc.

Moreover, as generative AI systems become more integrated into science workflows, watermarks provide a defense against hallucinated or manipulated data being misrepresented as trustworthy output. This is especially critical in domains like climate modeling, aerospace simulations, or biomedicine, where decisions based on tampered or unverifiable data can have far-reaching consequences.

From an enterprise perspective, watermarks serve three strategic functions:
\begin{enumerate}
    \item \textbf{Attribution and compliance:} Embedding identifiers into datasets helps ensure that AI systems comply with data licensing and attribution policies.
    \item \textbf{Trust and verification:} Watermarks enable downstream users - human or machine - to verify that data has not been altered during transmission, transformation, or inference.
    \item \textbf{Audit and traceability:} In multi-model workflows, watermarks allow forensic tracing of which systems or steps influenced the final output, supporting regulatory audit trails and intellectual property protection.
\end{enumerate}

As secure and explainable AI becomes a top priority for organizations deploying scientific and commercial ML systems, watermarking offers a practical, implementable solution. Our work demonstrates that physics-aware neural watermarking can bridge the gap between imperceptibility, decodability, and scientific fidelity - making it a viable tool for next-generation AI security strategies in science and industry.

\newpage

\section*{Acknowledgements}
\addcontentsline{toc}{section}{Acknowledgements}

We thank our colleagues and collaborators from the Massachusetts Institute of Technology, Arizona State University, Google DeepMind, Anthropic, and Lawrence Berkeley National Laboratory, and the University of California, Berkeley for their technical feedback throughout this project.

Special thanks to Kerri Lu for early-stage discussions and experimental feedback, and to John Kevin Cava for his support in physics-constrained modeling and data preparation. We are deeply grateful to Michael W. Mahoney for his mentorship and foundational guidance on scientific machine learning, security-aware systems design, and research framing.

This work benefited from an interdisciplinary exchange of ideas spanning neural watermarking, physics-informed learning, AI assurance, and enterprise-scale AI deployment. We are especially appreciative of the collaborative spirit that helped shape this research across institutional boundaries.

\newpage

\bibliographystyle{unsrt}
\bibliography{references}

\newpage

\appendix
\section*{Appendix}
\addcontentsline{toc}{section}{Appendix}

\subsection*{A.1 Training Diagnostics}

\begin{figure}[htbp]
    \centering
    \includegraphics[width=0.4\linewidth]{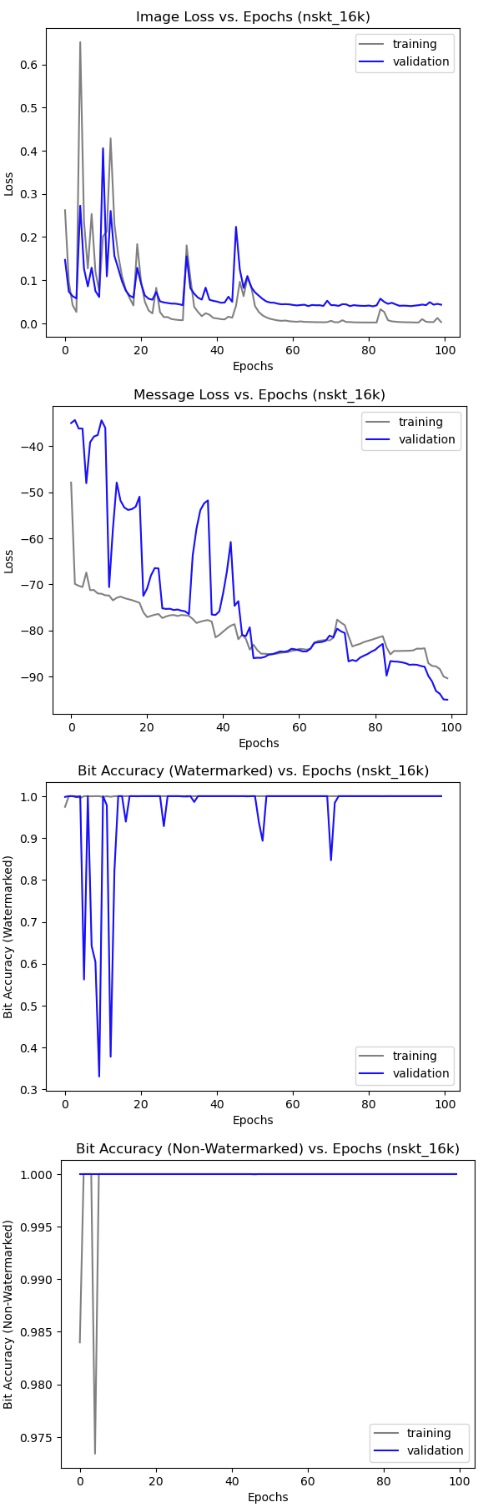}
    \caption{Training and validation loss curves for Fluid Flow UNet model.}
    \label{fig:appendix_fluid_loss}
\end{figure}

\begin{figure}[htp]
    \centering
    \includegraphics[width=0.45\linewidth]{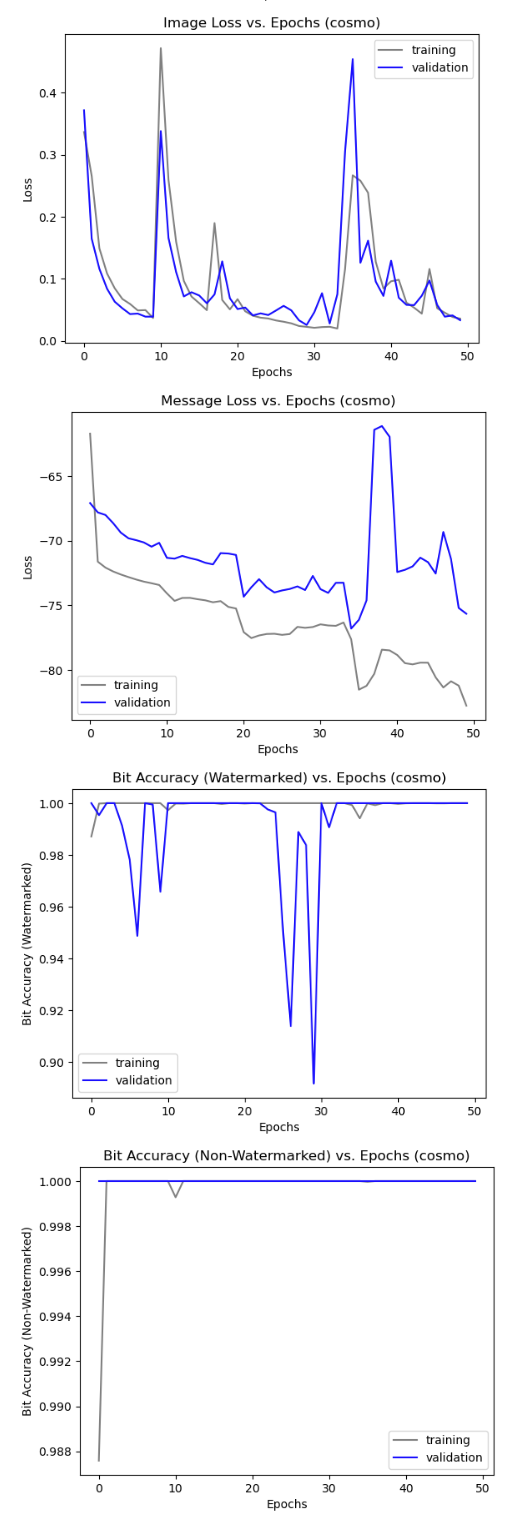}
    \caption{Training and validation loss curves for Cosmology UNet model.}
    \label{fig:appendix_cosmo_loss}
\end{figure}

Figures~\ref{fig:era5_loss}, \ref{fig:appendix_fluid_loss}, and \ref{fig:appendix_cosmo_loss} show the training and validation losses for the UNet model across the ERA5, Fluid Flow, and Cosmology datasets. For datasets with physics-aware training, physics loss is included alongside image and message loss.

\subsection*{A.2 Model and Training Configuration}

\begin{itemize}
    \item \textbf{Optimizer:} Adam  
    \item \textbf{Learning rate:} $1 \times 10^{-3}$  
    \item \textbf{Batch size:} 32  
    \item \textbf{Loss weights:} Equal weighting on image loss and message loss; physics loss weighted by $\lambda = 1.0$  
    \item \textbf{Early stopping:} Based on validation image loss and $\geq$98\% decoder accuracy  
    \item \textbf{Epochs:} Up to 100 pre-physics + 20 with physics loss (with early stopping)  
    \item \textbf{Hardware:} Trained on NERSC Perlmutter GPUs (A100, 40GB)
\end{itemize}

\subsection*{A.3 Source Code and Data Availability}

All code and experiment scripts used in this work are included in the following files:
\begin{itemize}
    \item \texttt{train\_UNet\_ERA5.ipynb}, \texttt{train\_UNet\_FluidFlow.ipynb}, \texttt{train\_UNet\_Cosmology.ipynb}
    \item \texttt{evaluate\_watermarkers.ipynb}
    \item Dataset loader: \texttt{SuperBench/src/data\_loader\_crop.py}
    \item UNet architecture: \texttt{SuperBench/unet\_utils/blocks.py}
\end{itemize}

\noindent\textbf{Code Repository:} \url{https://github.com/KrtiT/scientific-data-watermarker}

\subsection*{License and Usage Terms}

The source code for this project is made publicly available for non-commercial academic research purposes under the Creative Commons Attribution-NonCommercial 4.0 International (CC BY-NC 4.0) license.

\begin{itemize}
    \item Users are free to copy, distribute, and adapt the code for non-commercial purposes, provided appropriate credit is given to the original authors.
    \item Commercial use, including integration into proprietary systems or enterprise pipelines, is strictly prohibited without prior written permission.
    \item For commercial licensing inquiries, please contact the lead author or refer to the project GitHub repository for contact details and updates.
\end{itemize}

This license is effective as of \textbf{May 2025} and remains valid until superseded by a future update. Any modifications to the license will be documented in the GitHub repository's LICENSE file.

\end{document}